\documentclass{svproc}
\usepackage{url}
\usepackage{graphicx}
\usepackage{multicol}
\usepackage{footmisc}
\usepackage{hyperref}
\usepackage{multirow}
\usepackage{amsmath}
\usepackage{float}
\usepackage[normalem]{ulem}
\useunder{\uline}{\ul}{}

\begin{document}
\mainmatter             
\title{Homogeneous Feature Transfer and Heterogeneous Location Fine-tuning for Cross-City Property Appraisal Framework}
\titlerunning{Estate Valuation}  


\author{Yihan Guo\inst{1} \and Shan Lin\inst{1} \and Xiao Ma\inst{1} \and Jay Bal\inst{1} \and Chang-tsun Li\inst{1,2}}
\authorrunning{Yihan Guo et al.} 
%
\tocauthor{Yihan Guo, Shan Lin}
\institute{University of Warwick, Coventry, UK\\
\email{Yihan.Guo, Shan.Lin, X.Ma, Jay.Bal, C-T.Li@warwick.ac.uk}\\
Charles Sturt University, Waga Waga, Australia\\
\email{chli@csu.edu.au}}

\maketitle              

\begin{abstract}
Most existing real estate appraisal methods focus on building accuracy and reliable models from a given dataset but pay little attention to the extensibility of their trained model. As different cities usually contain a different set of location features (district names, apartment names), most existing mass appraisal methods have to train a new model from scratch for different cities or regions. As a result, these approaches require massive data collection for each city and the total training time for a multi-city property appraisal system will be extremely long. Besides, some small cities may not have enough data for training a robust appraisal model. To overcome these limitations, we develop a novel \textbf{H}omogeneous \textbf{F}eature \textbf{T}ransfer and \textbf{H}eterogeneous \textbf{L}ocation \textbf{F}ine-tuning \textbf{(HFT+HLF)} cross-city property appraisal framework. By transferring partial neural network learning from a source city and fine-tuning on the small amount of location information of a target city, our semi-supervised model can achieve similar or even superior performance compared to a fully supervised Artificial neural network (ANN) method.
\end{abstract}
\section{Introduction}
Real estate is always one of the pivotal parts of national economic development in many developing countries. Real estate is an enabler of business activity, that the growth of business activities requires consistent approaches to the valuation of real estate for accounting, banking activity, stock exchange listing and leverage lending purposes \cite{Mansfield2004ShapingGermany}. For these reasons, developing Automated Valuation Models(AVM) or Computer Assisted Mass Appraisal (CAMA) systems to support appraiser for more accurate property valuation is one crucial and recurrent research topics in both academia and industry. In the past three decades, many appraisal methodologies and applications have been proposed. They can be classified into two categories: traditional econometric approaches and machine learning approaches. The traditional approaches such as comparable method \cite{Pagourtzi2003RealMethods} and Multiple Regression Analysis (MRA)\cite{Benjamin2004MassValuation} usually rely on manual analysis of the characteristics of the properties. The recent machine learning based approaches cover a broad spectrum of machine learning techniques from k-Nearest Neighbors(k-NN)\cite{Chopra2007DiscoveringModel}, Regression Decision Trees\cite{Fan2006DeterminantsApproach}, Regression Random Forest\cite{Antipov2012MassDiagnostics}, Artificial Neural Networks(ANN)\cite{Bahrammirzaee2010ASystems}, rough sets theory\cite{Damato2007ComparingMethodologies} and some hybrid approaches\cite{Kontrimas2011TheIntelligence,Musa2013AValuation}.

One significant challenge for residential real estate appraisal modelling is to account for the differences in location. Some recently proposed methods introduce geographic information system (GIS) features into their models \cite{Chiarazzo2014ALocation,Musa2013AValuation,Fu2014ExploitingClustering}. However, most of these works only focus on analysing the geographic information within the same city but pay little attention to a cross-city scenario. When adding a new city into the AVM system, these approaches have to train a new model from scratch and require a massive data collection from the new city. Besides, the real estate markets in some low-tier cities are usually much smaller than those top-tier cities. Therefore, the amount of data collected from the small cities may not be sufficient enough to train a robust and reliable property appraisal model. If a robust model learned from one city can be transferred to other cities, the amount of time and resources required for training and data collection could be significantly reduced. Hence, in this paper, we proposed a novel \textbf{H}omogeneous \textbf{F}eature \textbf{T}ransfer and \textbf{H}eterogeneous \textbf{L}ocation \textbf{F}ine-tuning \textbf{(HFT+HLF)} cross-city property appraisal framework. As a case study of the Chinese real estate market, we collected the residential real estate resale data of six cites selected from three different city-tiers in China and evaluate the housing valuation performance of our methods after transfer learning. The contributions of our work are summarised below:
\begin{itemize}
\item Our work is one of the first research works focusing on cross-city transfer problem in property appraisal. Because the datasets collected from different cities contain entirely different sets of location features (district names, apartment names, etc.), most of the transfer learning methods cannot be directly applied to cross-city property appraisal models due to these heterogeneous location features. Hence, the cross-city property appraisal model transfer is a challenging task.\\
\item We proposed a novel \textbf{H}omogeneous \textbf{F}eature \textbf{T}ransfer and \textbf{H}eterogeneous \textbf{L}ocation \textbf{F}ine-tuning \textbf{(HFT+HLF)} framework. By transferring part of our neural network and semi-supervised fine-tuning the remaining part, our model can surpass the fully supervised single-city ANN model by using only 20\% to 30\% of the available training data.
\end{itemize}

\section{Related Work}
\subsection{Traditional Econometric Approaches} 
In the last few decades, there has been a large number of academic studies around the real estate appraisal area. Many of them employed the regression model for their valuation. Two major categories of traditional econometric approaches are hedonic regression models and hedonic price models. 

The hedonic regression models have been extensively researched in academia and widely used in the industry for residential real estate mass appraisal for the past three decades. They range from simple hedonic regression\cite{Isakson2001UsingAppraisal,Downes2002The1991}, ridge regression\cite{Ferreira1988RidgeAnalysis} to the more complex quantile regression\cite{Farmer2010UsingCompetition.,Narula2012ValuatingProgramming}. On the other hand, the hedonic price models make econometric analysis of the property attributes (e.g size, state, material etc.) as well as situational aspects such as the local environment (e.g. access bus and train stations) to understand trends in the housing market and accessing the factors which affect house prices. Additional variables such as inflation adjustment and regional planning strategy are often added \cite{Born1994RealCycles,Kanojia2016ValuationArt}. 

However, the traditional econometric models usually assume these house characteristic are independent and non-interrelated which means that the value influence of the property attributes is considered to be constant. This assumption usually cannot correctly reflect the real-world real estate market. The traditional econometric methods are essentially model-oriented approaches which are aiming to explain the real estate prices and their variations.



\subsection{Machine Leaning Based Approaches}
The recent development of machine learning and deep learning are primarily driven by the abundance of available data and advances in computer technology. Many research works now focus on implementing the machine learning techniques into real estate appraisal system. Unlike the traditional approaches which are modelled assuming explicit rules, the machine learning based approaches try to learning the feature to price mapping automatically from data. 

The genetic algorithm (GA) is a machine learning method which has been introduced into real estate appraisal in recent year \cite{DelGiudice2017UsingAppraisals,Kauko2003ResidentialApproaches}. The GA based approaches consider the real estate appraisal task as a multi-parameter optimisation problem and tackle this problem by stochastic search techniques based on Charles Darwin's evolutionary principle \cite{Goldberg1988GeneticLearning}. The decision tree approaches had also been applied to real estate appraisal area to overcome the potential problems relating to fundamental model assumption and independent variables \cite{Fan2006DeterminantsApproach}. Easy to understand and the ability to handle categorical variables make the decision tree approaches frequently used in many property appraisal systems. The random forest approach is an extended decision tree method by ensemble many simple regression trees to increase the overall appraisal accuracy \cite{Antipov2012MassDiagnostics}. For small and noise-free dataset, the random forest method is one of the most accurate approaches comparing with many other approaches like KNN or ANN \cite{Antipov2012MassDiagnostics}. However, in a noisy real-world dataset, the performance of the random forest approach is significantly worse than ANN-based models \cite{Kempa2011InvestigationAppraisal}. Overlooking all kinds of models, the most commonly used machine learning approaches in the recent year are Artificial Neural Networks(ANN) \cite{Chiarazzo2014ALocation,Bahrammirzaee2010ASystems,Nghiep2001PredictingNetworks,Mccluskey1999TheProperties}. The ANN-based approaches can outperform most of the traditional methods if the training dataset is large enough and the right training parameters are set \cite{Nghiep2001PredictingNetworks}. Because the ANN approach yields the best performance when handling large datasets, our cross-city transfer model is developed base on ANN structure as well.  

Most of the machine learning based studies only consider the location as a primary feature variable for training real estate appraisal models. However none has paid attention to location extensibility of the property appraisal models. For each new location such as a new city or a new region, a new model has to be retrained from scratch in order to accommodate the different location information. To overcome the particular limitation, we propose a cross-city transferable ANN-based property appraisal framework.

\section{Methodology}
\subsection{Homogeneous and Heterogeneous Features}
\begin{figure}[h]
\centering
\includegraphics[trim=1cm 0.6cm 1cm 0.6cm,clip,width=1\textwidth]{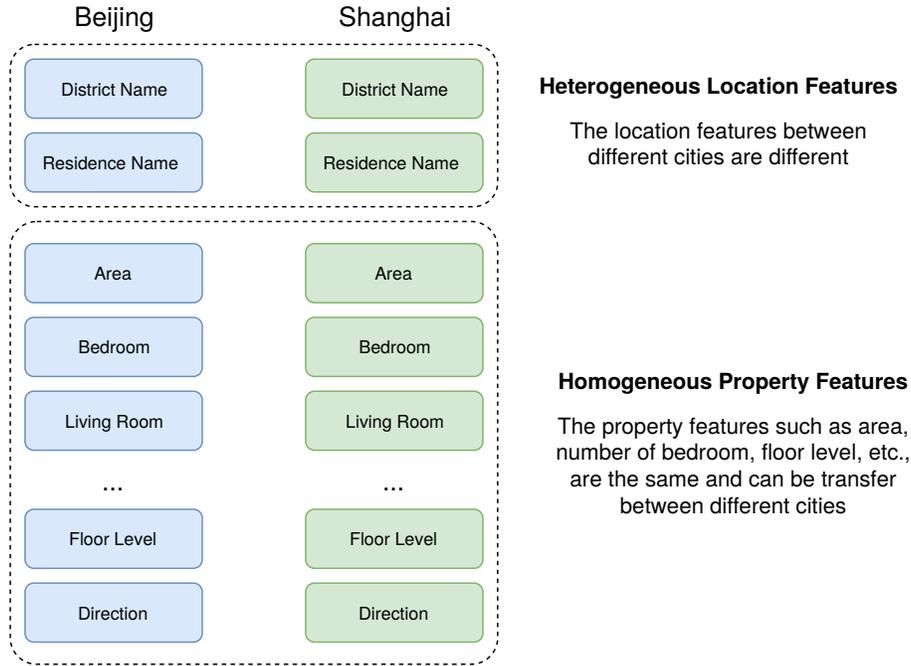}
\caption{The cross-city estate dataset can be divided into two categories: homogeneous features and heterogeneous features. The homogeneous features usually describe the characteristics of the property such as direction, floor level, area. These features are usually city invariant. However, the location information like district names and apartment community names are city-dependent}
\label{fig:data}
\end{figure}

Real estate datasets usually contain three feature categories: location characteristics, building characteristics and apartment characteristics. The location characteristics tell where the buildings are located. The building characteristics describe the condition of the entire building and its neighbourhood area. The apartment characteristics are the apartment's internal construction such as the number of bedroom or living room. The variables of building characteristics and apartment characteristics are usually homogeneous for different cities. For example, the decoration level of two apartments of city A and city B can both be labelled from the same set of categorical variables: None, Partial, Simple, Mid-range, Deluxe, or Luxury. However, different cities have different sets of location information: different districts, different residential communities, etc. Hence, the location variables are heterogeneous for different cities. The heterogeneous location features and homogeneous property features are demonstrated in Fig \ref{fig:data}.

\subsection{HFT+HLF Framework}



\begin{figure}[H]
\centering
\includegraphics[trim=0cm 0cm 0cm 0cm,clip,width=0.8\textwidth]{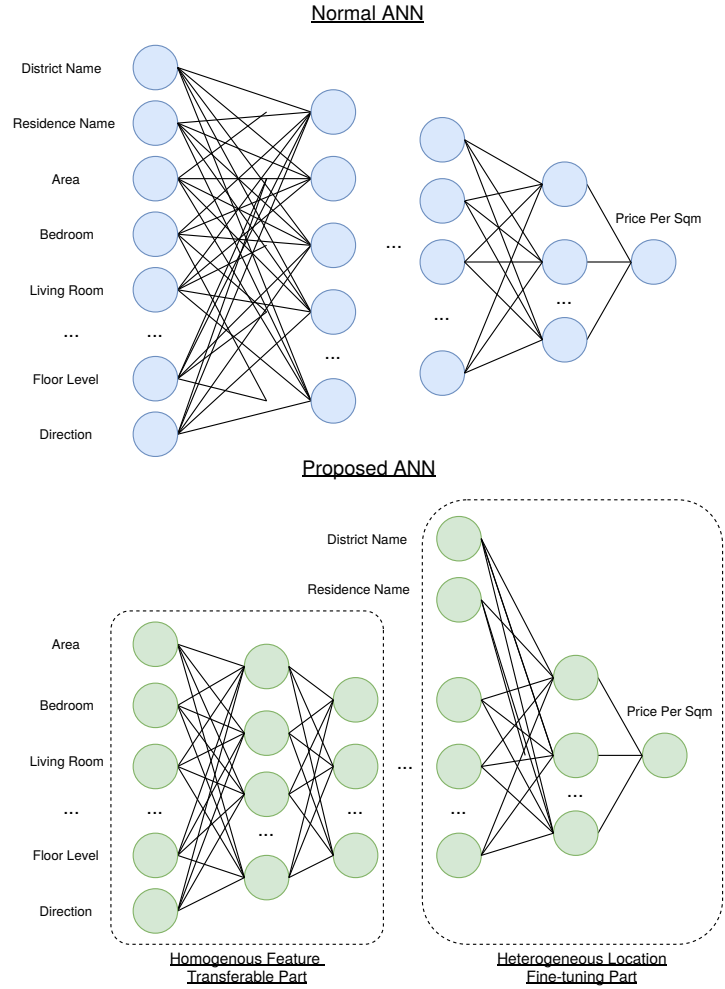}
\vspace{-2em}
\caption{The top diagram is the traditional ANN structure for property appraisal. The bottom figure is our proposed structure tailored for cross-city transfer learning. The transferable part can be transferred to the new model of a different city. Only the parameters in the fine-tune part need to be learned based on the data from a new city}
\label{fig:architecture}
\end{figure}

Most of the ANN-based property appraisal methods have to replace and retrain the regressors for each new city due to the heterogeneous location features. Our model addresses this problem by separating the homogeneous and heterogeneous features during the training. The homogeneous features will be used to learn a cross-city transferable model and the city-dependant heterogeneous features will be used for location-based fine-tuning. As a result, our proposed network consists of two joined ANNs: homogeneous feature transferable ANN and heterogeneous location fine-tuning ANN, as shown in Fig \ref{fig:architecture}. The homogeneous features such as apartment features and building features will be the inputs for the transferable part. The output feature maps from the transferable ANNs will then be concatenated with the heterogeneous location features and become the new inputs to the fine-tuning section of our proposed network. Since the homogeneous features are commonly shared between apartments of different cities, this part of our neural network can be considered as a generic apartment and building features learning network which could be transferred between different cities. By transfer this part of our network to a new city, the new city model only need to optimise the weight of the fine-tuning network. For example, the weights of the transferable ANN learned from the source city Beijing can be transferred to the new model of the target city Baotou and then fine-tuned with only a few data from the target city Baotou. The transferable ANN consists of 5 hidden layers with $(200,100,50,20,10)$ hidden nodes and the fine-tuning ANN consists of 4 hidden layers with $(100,50,20,10)$ hidden nodes. All the hidden layers are equipped with 0.1 negative slope Leaky ReLU activation function, a 0.5 drop-out rate and Batch Normalization. We use the popular mean squared error (MSE) as the criterion loss for the price regression. Our network can be optimised by Adam algorithm with AMSGrad moving average variant. The overview of our architecture is shown in the lower section of Fig \ref{fig:architecture} with the standard traditional ANN structure added on top for comparison.
\vspace{-3em}

\section{Empirical Evaluation}
\subsection{Database}
In addition to the proposed location transfer appraisal method, another contribution of our study is the introduction and employment of a new massive cross-city dataset. The dataset was formed as a result of the collaboration between 2 partners, JinZheng and Cityre.  JinZheng real estate appraisal company is a top level real estate appraisal Company in China, which cover all regions in China. Also, Cityre is a leading company of national real estate property information and data-service provider in China. This dataset contains two cities for each of three city tiers of China: Tier 1 includes 127,441 raw sales records (sale tag price) for Beijing, 243,222 raw records for Shanghai; Tier 2 includes 47,124 raw records for Jinan and 67,611 for Qingdao; Tier 3 includes that 5,609 raw records for Hohehaot, 9,614 raw records for Baotou. These cities were selected from the national setting of China for city tier classification. Beijing and Shanghai are the most famous Tier 1 city in China. Jinan and Qingdao are both Tier 2 cities from the same province. Hohhot and Baotou are Tier 3 cities from the same province. The details of our dataset is demonstrated in Table \ref{table:data_detail}:
\vspace{-1em}
\begin{table}[h]
\begin{tabular}{c|c|c|c|c|c|c|c}
\hline
Tier & \multicolumn{2}{c|}{Tier 1} & \multicolumn{2}{c|}{Tier 2} & \multicolumn{2}{c|}{Tier 3} & \multirow{2}{*}{Total} \\ \cline{1-7}
City & Beijing & Shanghai & Jinan & Qingdao & Hohhot & Baotou &  \\ \hline
No. of District & 18 & 18 & 9 & 12 & 5 & 4 & 66 \\
No. of Residence & 1,241 & 2,581 & 443 & 580 & 84 & 110 & 4,997 \\
No. of Raw Data & 278,371 & 580,211 & 104,149 & 133,988 & 19,334 & 25,407 & 1,141,460 \\
\hline
No. of Processed Data & 127,441 & 243,222 & 47,124 & 67,611 & 5,609 & 9,614 & 500,621 \\ \hline
\end{tabular}
\caption{The dataset is divided into three city tiers. Each city-tier contains two most famous or economically impactful cities. The raw records contain many missing values. The final processed data is the clean up data used for the empirical evaluation}
\label{table:data_detail}
\end{table}
\vspace{-2em}

Our dataset surpasses most of the datasets used in most of the studies of property valuation. Brown and D'amato's work only use 725 and 390 dwellings receptively \cite{Brown2004APrices,Damato2007ComparingMethodologies}. The recent works from García and Arribas only use 591 and 2,149 records \cite{Cervello2011RankingModel,Arribas2016MassModelling}. All six cities in our dataset have more data compared to all of them. Each residential real estate record in our dataset consisting of 15 variables were collected for each apartment: 3 describe the location, 10 describe the characteristics, 2 describe the building in which it is sited. The detailed descriptions of the variables are the following:\\
\vspace{-2em}
\begin{itemize}
\item \textbf{Location Characteristics}
\begin{enumerate}
\item \underline{city}: Six cities (Beijing, Shanghai, Jinan, Qingdao, Hohhot, Baotou).
\item \underline{district}: District the apartment is located
\item \underline{residence}: Name of the residential apartment
\end{enumerate}
\item \textbf{Building Characteristics}
  \begin{enumerate}
  \item \underline{year}: Year the building is completed
    \begin{itemize}
  	\item Mean 2003.95, Median 2005, Min 1900, Max 2019, Std 7.90
  	\end{itemize}
  \item \underline{building\_type}: 8 building types (Bungalow, High-rise, High-level, Multi-storey, Entire Block, Semi-detached House, Detached House, Siheyuan)
  \end{enumerate}
\item \textbf{Apartment Characteristics}
\begin{enumerate}
\item \underline{price}: Price per square meter
  \begin{itemize}
  \item Mean 46,878, Median 44,666, Min 2,075, Max 288,690, Std 28,810
  \end{itemize}
\item \underline{area}: Total area of the apartment
  \begin{itemize}
  \item Mean 99.89, Median 88, Min 10, Max 2900, Std 59.87
  \end{itemize}
\item \underline{bedroom}: Number of bedrooms
  \begin{itemize}
  \item Mean 2.29, Median 2, Min 0, Max 9, Std 1.03
  \end{itemize}
\item \underline{livingroom}: Number of livingrooms
  \begin{itemize}
  \item Mean 1.56, Median 2, Min 0, Max 7, Std 0.56
  \end{itemize}
\item \underline{kitchen}:Number of kitchen
  \begin{itemize}
  \item Mean 0.97, Median 1, Min 0, Max 5, Std 0.19
  \end{itemize}
\item \underline{bathroom}: Number of bathroom
  \begin{itemize}
  \item Mean 1.32, Median 1, Min 0, Max 9, Std 0.64
  \end{itemize}
\item \underline{floor}: Floor number of the apartment
  \begin{itemize}
  \item Mean 5.36, Median 4, Min -10, Max 63, Std 4.94
  \end{itemize}
\item \underline{structure}: Apartment Structure.
\item \underline{decoration}: 6 levels of the decoration (None, Partial, Simple, Mid-range, Deluxe, Luxury)
\item \underline{direction}: 10 Direction of the property (North, South, East, West, NorthEast, NorthWest, SouthEast, SouthWest, NorthSouth, EastWest)
\end{enumerate}
\end{itemize}
\vspace{-1em}

\subsection{Training and Settings}
Cities of different tiers normally have different sizes of the training samples available. As shown in Table \ref{table:data_detail}, the available data of first-tier cities (Beijing, Shanghai) is nearly 20 times more than the third-tier cities (Hohhot, Baotou). As a result, the batch size and the learning rate have to be individually set for each city tier. In our experiment, we set the learning rate at 0.005 with batch size 256 for Tier 1 cities, a 0.01 learning rate with batch size 128 for Tier 2 cities and 0.02 learning rate with batch size 64 for Tier 3 cities. The number of epochs is set 250 to ensure our network is fully converged during the training. Our network is implemented in PyTorch and the training times range from 2 hours to 30 minutes depending on the training sample size. We adopted the commonly used Root Mean Square Error (RMSE), R-squared Error ($R^2$) and Mean Absolute Percentage Error (MAPE) as performance metrics.

\subsection{Experiment 1: Traditional ANN vs Our Proposed ANN}
Since our proposed network is a new ANN structure, the first experiment is to check whether the new structure affects the appraisal performance of the ANN model. In this experiment, we compare the performance of our proposed transferable model with a normal non-transferable ANN model in a fully supervised single city setting. This experiment follows the common 10-fold cross-validation strategy which is the average of the experimental results based on 10 random splits of the dataset into 90\% training samples and 10\% testing samples. The detailed performance metrics are shown in Table \ref{table:experiment1} below.
\vspace{-1em}
\begin{table}[H]
\centering
\resizebox{\textwidth}{!}{%
\begin{tabular}{c|c|cc|cc|cc}
\hline
\multirow{2}{*}{Model} & \multirow{2}{*}{Performance Metric} & \multicolumn{2}{c|}{Tier 1} & \multicolumn{2}{c|}{Tier 2} & \multicolumn{2}{c}{Tier 3} \\
 &  & Beijing & Shanghai & Jinan & Qingdao & Hohhot & Baoto \\ \hline
\multirow{3}{*}{Traditional ANN} & RMSE (Lower Better) & 0.246 & 0.186 & \textbf{0.184} & \textbf{0.202} & \textbf{0.150} & 0.145 \\
 & MAPE (Lower Better)& 0.151 & 0.138 & \textbf{0.142} & 0.157 & \textbf{0.121} & 0.108 \\ 
 & $R^2$ (Higher Better) & 0.804 & \textbf{0.782} & \textbf{0.757} & \textbf{0.828} & 0.522 & \textbf{0.485} \\ \hline
\multirow{3}{*}{Proposed ANN} & RMSE (Lower Better) & \textbf{0.239} & \textbf{0.185} & 0.189 & 0.204 & \ 0.170 & \textbf{0.135} \\
 & MAPE (Lower Better) & \textbf{0.149} & \textbf{0.136}  & 0.147 & 0.159 & 0.129 & \textbf{0.102} \\ 
 & $R^2$ (Higher Better) & \textbf{0.816} & 0.778 & 0.744 & \textbf{0.828} & \textbf{0.523} & 0.473 \\ \hline
\end{tabular}
}
\caption{Fully Supervised Single City Learning Performance Comparison between Traditional ANN and Our Transferable ANN. The best results are highlighted by Bold}
\label{table:experiment1}
\end{table}
\vspace{-2em}

Our model achieved similar and impressive low RMSE and MAPE scores in all six cities which means that our model can converge well on all the training datasets. However, the low $R^2$ scores of the Tier 3 city indicate the poor regression performance. It is mainly due to the lack of available data in the Tier 3 cities. This discovery supports our claim that the cross-city transfer learning is necessary for property appraisal model, especially to those low volume Tier 3 cities. By comparing our proposed models with standard non-transferable ANN method, our network yields a very similar overall performance. It proves that the new proposed architecture did not affect the overall appraisal performance.

\subsection{Experiment 2: Semi-supervised Cross City Transfer Learning}
Most other ANN appraisal methods, our proposed model has the ability to transfer a partially pre-trained network learned from the source city to the target city. Experiment 2 is conducted to validate the performance of our model in a cross-city semi-supervised setting. 
\vspace{-1em}
\subsubsection{Transfer to Tier 1 Cities:}
The first set of experiments is to test our model's ability to transfer the pre-train network to the first tier cities: Beijing and Shanghai. Based on the situation of China, the Tier 1 cities usually have a massive amount of data available for training. Therefore, the demand for transferring appraisal models conducted form Tier 2 or 3 cities to Tier 1 cities is highly unlikely in real-world practice. As a result, we conduct the Beijing to Shanghai and Shanghai to Beijing Transfer Learning in this experiment. For Shanghai to Beijing transfer, we first pre-train our model based on the training dataset of Shanghai and transfer the transferable part of our network to the new model for Beijing. Then, we select 10, 20, 30 records from each location (each residential community) in Beijing to fine tune the Beijing model. The test dataset is randomly selected from the remaining dataset with the size of 10\% of the overall Beijing dataset. The processes for Beijing to Shanghai transfer are the same. As we only use a small amount of data from the Beijing training dataset, it can be considered as semi-supervised learning. The detailed performance metrics are shown in Table \ref{table:tier1}.

\vspace{-2em}
\begin{table}[H]
\centering
\resizebox{\textwidth}{!}{%
\begin{tabular}{c|c|c|c|c}
\hline
\multirow{2}{*}{Semi Supervised} & \multirow{2}{*}{\begin{tabular}[c]{@{}c@{}}Training\\ Size\end{tabular}} & Beijing ($R^2$ Score) & \multirow{2}{*}{\begin{tabular}[c]{@{}c@{}}Training\\ Size\end{tabular}} & Shanghai ($R^2$ Score) \\ \cline{3-3} \cline{5-5} 
 &  & Shanghai -\textgreater Beijing &  & Beijing -\textgreater Shanghai \\ \hline
\multicolumn{1}{l|}{10 Records Per Residence} & 12,020 & 0.829 & 25,630 & 0.758 \\
20 Records Per Residence & 24,040 & \textbf{0.847} & 50,860 & 0.776 \\
30 Records Per Residence & 36,060 & \textbf{0.845} & 76,290 & \textbf{0.788} \\ \hline
\begin{tabular}[c]{@{}c@{}}Full Supervised\\ (No Transfer Learning)\end{tabular} & 127,441 & {\ul 0.816} & 243,222 & {\ul 0.778} \\ \hline
\end{tabular}%
}
\caption{The $R^2$ scores for Shanghai to Beijing and Beijing to Shanghai Transfer Learning.}
\label{table:tier1}
\end{table}
\vspace{-3em}

By only using 20 records from each residential community, our semi-supervised transferable model can quickly achieve the similar or even superior performance compared with fully supervised single city learning. If the training dataset increase to 30 records per residential community, the overall performance even surpass the single city supervised learning. If the training set is 20 records from each residential community, the size of the training data is only one-fifth of the original training dataset. As a result, our proposed transferable model can be significantly reduced by five times.
\vspace{-1em}

\subsubsection{Transfer to Tier 2 Cities}
Table \ref{table:tier2} shows the experiment results for Tier 1 Cities to Tier2 Cities Transfer and Within Tier 2 Cities Transfer. The Tier 1 to Tier 2 transfer models usually outperform the inter Tier 2 transfer model. Because the Tier 1 cities have much more training data than Tier 2 and 3 cities, the transferable part of our network learned from Tier 1 cities usually have better generalization-ability and robustness.
\vspace{-1em}
\begin{table}[H]
\centering
\resizebox{\textwidth}{!}{%
\begin{tabular}{c|c|c|c|c|c|c|c|c}
\hline
\multirow{2}{*}{Semi-Supervised} & \multirow{2}{*}{\begin{tabular}[c]{@{}c@{}}Training \\ Size\end{tabular}} & \multicolumn{3}{c|}{Jinan ($R^2$ Score)} & \multirow{2}{*}{\begin{tabular}[c]{@{}c@{}}Training \\ Size\end{tabular}} & \multicolumn{3}{c|}{Qingdao ($R^2$ Score)} \\ \cline{3-5} \cline{7-9} 
 &  & Beijing -\textgreater Jinan & Shanghai -\textgreater Jinan & Qingdao -\textgreater Jinan &  & Beijing -\textgreater Qingdao & Shanghai -\textgreater Qingdao & Jinan -\textgreater Qingdao \\ \hline
10 Records Per Residenc & 4,400 & 0.558 & 0.546 & 0.495 & 5,790 & 0.624 & 0.636 & 0.608 \\
20 Records Per Residence & 8,800 & 0.642 & 0.612 & 0.561 & 11,580 & 0.756 & 0.755 & 0.659 \\
30 Records Per Residence & 13,200 & \textbf{0.784} & \textbf{0.773} & 0.614 & 17,370 & \textbf{0.836} & \textbf{0.838} & 0.732 \\ \hline
\begin{tabular}[c]{@{}c@{}}Full Supervised\\ (No Transfer Learning)\end{tabular} & 47,124 & \multicolumn{3}{c|}{0.744} & 67,611 & \multicolumn{3}{c|}{0.828} \\ \hline
\end{tabular}%
}
\caption{The $R^2$ scores for Tier 1 to Tier 2 Transfer and Inter-Tier-2 Transfer}
\label{table:tier2}
\end{table}
\vspace{-4em}

\subsubsection{Transfer to Tier 3 Cities}
As the Tier 3 cities have limited amount of training data and the supervised Tier 3 cities models have relatively poor performance (very low $R^2$ scores), our experiment did not include the Tier 3 to Tier 3 cities transfer evaluation. We only focus on evaluating our network on Tier 1 or Tier 2 cities transfer to Tier 3.
\vspace{-1em}
\begin{table}[H]
\centering
\resizebox{\textwidth}{!}{%
\begin{tabular}{c|c|c|c|c|c|c}
\hline
\multirow{2}{*}{Semi-Supervised} & \multirow{2}{*}{\begin{tabular}[c]{@{}c@{}}Training \\ Size\end{tabular}} & \multicolumn{2}{c|}{Hohhot ($R^2$ Score)} & \multirow{2}{*}{\begin{tabular}[c]{@{}c@{}}Training \\ Size\end{tabular}} & \multicolumn{2}{c|}{Baotou ($R^2$ Score)} \\ \cline{3-4} \cline{6-7} 
 &  & Beijing -\textgreater Hohhot & Shanghai -\textgreater Hohhot &  & Beijing -\textgreater Baotou & Shanghai -\textgreater Baotou \\ \hline
5 Records Per Residence & 420 & 0.459 & 0.446 & 550 & 0.424 & 0.436 \\
10 Records Per Residence & 840 & 0.486 & 0.512 & 1100 & 0.456 & 0.455 \\
15 Records Per Residence & 1260 & \textbf{0.547} & \textbf{0.588} & 1650 & \textbf{0.513} & \textbf{0.520} \\ \hline
\begin{tabular}[c]{@{}c@{}}Full Supervised\\ (No Transfer Learning)\end{tabular} & 5048 & \multicolumn{2}{c|}{0.523} & 8652 & \multicolumn{2}{c|}{0.473} \\ \hline
\end{tabular}%
}
\caption{The $R^2$ scores for Tier 1 to Tier 3 Transfer}
\label{table:tier3-1}
\end{table}
\vspace{-1em}

\vspace{-2em}
\begin{table}[H]
\centering
\resizebox{\textwidth}{!}{%
\begin{tabular}{c|c|c|c|c|c|c}
\hline
\multirow{2}{*}{Semi Supervised} & \multirow{2}{*}{\begin{tabular}[c]{@{}c@{}}Training\\ Size\end{tabular}} & \multicolumn{2}{c|}{Hohhot ($R^2$ Score)} & \multirow{2}{*}{\begin{tabular}[c]{@{}c@{}}Training\\ Size\end{tabular}} & \multicolumn{2}{c|}{Baotou ($R^2$ Score)} \\ \cline{3-4} \cline{6-7} 
 &  & Jinan -\textgreater Hohhot & Qingdao -\textgreater Hohhot &  & Jinan -\textgreater Baotou & Qingdao -\textgreater Baotou \\ \hline
5 Records Per Residence & 420 & 0.509 & 0.482 & 550 & 0.347 & 0.422 \\
10 Records Per Residence & 840 & 0.521 & 0.555 & 1100 & 0.448 & 0.451 \\
15 Records Per Residence & 1260 & \textbf{0.548} & \textbf{0.585} & 1650 & \textbf{0.516} & \textbf{0.521} \\ \hline
\begin{tabular}[c]{@{}c@{}}Full Supervised\\ (No Transfer Learning)\end{tabular} & 5048 & \multicolumn{2}{c|}{{\ul 0.523}} & 8652 & \multicolumn{2}{c|}{{\ul 0.473}} \\ \hline
\end{tabular}%
}
\caption{The $R^2$ scores for Tier 2 to Tier 3 Transfer}
\label{table:tier3-2}
\end{table}
\vspace{-3em}

Table \ref{table:tier3-1} demonstrate the model transfer from Tier 1 cities (Beijing, Shanghai) to Tier 3 cities (Hohhot, Baotou). Table \ref{table:tier3-2} demonstrate the model transfer from Tier 2 cities (Jinan, Qingdao) to Tier 3 cities (Hohhot, Baotou). By using only 15 records from each residential community, our model already outperforms the fully supervised model. In other words, the performance of our proposed model yield significant improvement after transferring either Tier 1 to Tier 3 or Tier 2 to Tier 3 cities. It shows that by transferring the homogeneous property features learning network trained from a more substantial training data (Tier 1 or 2 cities) can help to boost the performance of low data volume Tier 3 cities. 

\section{Conclusion}
In this paper, we focus on solving a challenging problem in most of the existing property appraisal models: lack of adaptiveness and extensiveness for a different location. Because data for different cities contain completely different location feature, traditional property appraisal models have to be trained from scratch for each city. To address this problem, we presented a novel semi-supervised homogeneous features transfer and heterogeneous location fine-tuning network. We reconstruct the artificial neural network (ANN) into transferable homogeneous feature learning and heterogeneous location fine-tuning. By transferring the homogeneous feature learning component from a source city and fine-tune by a small amount of the target city's location feature, our semi-supervised model can achieve a similar or even superior performance of a fully supervised model. Meanwhile, the amount of training data required for our model is only 20\% of fully supervised ANN models. By this proposed method, real estate appraisal models trained on data-rich cities can be applied to cities with insufficient real estate data without compromising the accuracy. It could be used to reduce data collection period, lower the model training cost and establish a better economy benchmark.

\bibliography{reference}
\bibliographystyle{spmpsci}
\end{document}